\documentclass{article}

\usepackage{microtype}
\usepackage{graphicx}
\usepackage{subfigure}
\usepackage{booktabs} 

\usepackage{hyperref}

\usepackage[accepted]{icml2024}

\usepackage{amsmath}
\usepackage{amssymb}
\usepackage{mathtools}
\usepackage{amsthm}

\usepackage[capitalize,noabbrev]{cleveref}

\theoremstyle{plain}

\theoremstyle{definition}

\theoremstyle{remark}

\usepackage[textsize=tiny]{todonotes}

\newcommand{\out}[1]{}
\usepackage{algorithm}
\usepackage{algorithmic}

\usepackage{amsmath,amsfonts,bm}

\def\eqref#1{equation~\ref{#1}}

\def\floor#1{\lfloor #1 \rfloor}
\def\1{\bm{1}}

\DeclareMathAlphabet{\mathsfit}{\encodingdefault}{\sfdefault}{m}{sl}
\SetMathAlphabet{\mathsfit}{bold}{\encodingdefault}{\sfdefault}{bx}{n}

\usepackage{booktabs}
\usepackage{amssymb,amsmath}

\newcommand{\ours}{\texttt{MFRNP}}

\newcommand{\x}{x}

\icmltitlerunning{Multi-Fidelity Residual Neural Processes}

\begin{document}

\twocolumn[

\icmltitle{Multi-Fidelity Residual Neural Processes for Scalable Surrogate Modeling}

\icmlsetsymbol{equal}{*}

\begin{icmlauthorlist}
\icmlauthor{Ruijia Niu}{cse}
\icmlauthor{Dongxia Wu}{cse}
\icmlauthor{Kai Kim}{cse}
\icmlauthor{Yi-An Ma}{ds}
\icmlauthor{Duncan Watson-Parris}{ds,scripps}
\icmlauthor{Rose Yu}{cse}
\end{icmlauthorlist}

\icmlaffiliation{cse}{Department of Computer Science and Engineering, University of California San Diego, La Jolla, California, USA}
\icmlaffiliation{ds}{Halıcıoğlu Data Science Institute, University of California San Diego, La Jolla, California, USA}
\icmlaffiliation{scripps}{Scripps Institution of Oceanography, University of California San Diego, La Jolla, California, USA}

\icmlcorrespondingauthor{Rose Yu}{roseyu@ucsd.edu}

\icmlkeywords{Machine Learning, ICML}

\vskip 0.3in
]



\printAffiliationsAndNotice{} 

\begin{abstract}
Multi-fidelity surrogate modeling aims to learn an accurate surrogate at the highest fidelity level by combining data from multiple sources. Traditional methods relying on Gaussian processes can hardly scale to high-dimensional data. Deep learning approaches utilize neural network based encoders and decoders to improve scalability. These approaches share encoded representations across fidelities without including corresponding decoder parameters. This hinders inference performance, especially in out-of-distribution scenarios when the highest fidelity data has limited domain coverage. To address these limitations, we propose Multi-fidelity Residual Neural Processes (\ours{}), a novel multi-fidelity surrogate modeling framework. \ours{} explicitly models the residual between the aggregated output from lower fidelities and ground truth at the highest fidelity. The aggregation introduces decoders into the information sharing step and optimizes lower fidelity decoders to accurately capture both in-fidelity and cross-fidelity information. We show that \ours{} significantly outperforms  state-of-the-art in learning partial differential equations and a real-world climate modeling task. Our code is published at: \href{https://github.com/Rose-STL-Lab/MFRNP}{\texttt{github.com/Rose-STL-Lab/MFRNP}}.

\end{abstract}

\section{Introduction}
From engineering to climate science, a computational model, often realized by simulation, is frequently used to characterize the input-output relationship of a physical system. These computational models can be simulated at multiple levels of sophistication. The high-fidelity simulators are more accurate but resource demanding, whereas lower-fidelity models are less accurate but more computationally efficient. For example, in climate science, people incorporate real-world observations with computational models to calibrate simulations \cite{hersbach2020era5, karami2020smart}. While the calibrated data has refined details and accuracy, the domain coverage is very limited and the simulation process is computation heavy. Multi-fidelity surrogate modeling \cite{peherstorfer2018survey} aims to balance the computation efficiency-accuracy trade-off by utilizing data across fidelities to learn an accurate surrogate at the highest fidelity. 

Gaussian Processes (GPs) \cite{seeger2004gaussian} are a popular tool for surrogate modeling. Recent works have attempted to extend GP to multi-fidelity setting \citep{perdikaris2016multifidelity,wang2021multi}. However, they inherit the limited scalability from GP due to the inversion of the kernel matrix \citep{williams1995gaussian, rasmussen2003gaussian}. To solve this issue, many have proposed deep learning-based surrogate models \citep{damianou2013deep, raissi2016deep, salimbeni2017doubly, wilson2016deep}. Neural Processes (NPs) \citep{wang2020mfpc,hebbal2021multi} stand out as one of the most appealing approaches regarding inference performance and scalability. NPs are capable of encoding fidelity-specific data into low-dimensional latent representations and use them to improve inference performance at the highest fidelity, alleviating the scalability issue from high-dimensional data.

For example, Deep Multi-fidelity Active Learning (DMFAL) \cite{li2022deep, li2022batch} proposed to pass information from lower fidelities to higher fidelities with encoded hidden representations. This method requires a hierarchical structure in the latent space passing information from low to high fidelity level, which can lead to the error propagation issue. Disentangled Multi-fidelity Deep Bayesian Active Learning (D-MFDAL) \cite{wu2023disentangled} alleviates this issue by redesigning the NP using local and global hidden representations. Nevertheless, these methods rely on latent representations from only the encoders for cross-fidelity information sharing. However, the decoder parameters varies across fidelities and are not shared. At the highest fidelity, shared representations are decoded with different parameters, making the decoded output inherently inaccurate. This significantly limits the inference performance, especially when the model needs to extrapolate with  out-of-distribution (OOD) inputs when the training data has limited domain coverage at the highest fidelity.

In this work, we introduce a novel multi-fidelity surrogate modeling framework, Multi-fidelity Residual Neural Process (\ours{}), to address the aforementioned issues. \ours{} aggregates the predictions from surrogate models across lower-fidelity levels and employs an NP surrogate to capture the residual between the aggregated prediction and the ground truth at the highest fidelity level. By directly utilizing the outputs from lower-fidelity surrogates to share information, \ours{} includes decoders in cross-fidelity information sharing, improving accuracy while maintaining scalability. Moreover, we developed a tailored Evidence Lower Bound, named Residual-ELBO, to serve as our loss function. This novel loss function ensures the highest fidelity latent variable $z_K$ depends on all the other latent variables and decoder parameters across fidelities. Thus, \ours{} ensures accurate information aggregation from lower fidelities to promote residual modeling at the highest fidelity. To summarize, our contributions include:

\begin{itemize}
    \item A novel multi-fidelity surrogate model, Multi-fidelity Residual Neural Process (\ours{}). Its architecture shares input-specific information from lower fidelities, tackles the varying decoder problem with no error propagation yet preserving scalability.

    \item A novel Residual-ELBO to simultaneously promote learning across fidelities  and optimize lower fidelity decoders for residual modeling at the highest fidelity.

    \item Superior performance in global scale real-world climate modeling and numerous benchmark tasks on partial differential equations. \ours{} outperform the state-of-the-art baseline by \textbf{$\sim$90\%} in average.

\end{itemize}

\section{Background}
\paragraph{Multi-Fidelity Modeling.} 
Multi-fidelity modeling aims to capture the complex mapping from low-dimensional input variables $\mathcal{X} \subseteq  \mathbb{R}^{d_x}$ to high-dimensional output $\mathcal{Y} \subseteq  \mathbb{R}^{d_y}$ of the function $f:\mathcal{X} \rightarrow \mathcal{Y}$. For systems with $K$ fidelities where $K>1$, the cost $c_k$ of evaluating $f_k\in \{f_{1},\cdots, f_{K}\}$ increases with the fidelity level $(c_1 <\cdots < c_K )$ as $f_k$ conveying more detailed information in approximating $f$. Our goal is to learn a deep surrogate model  $\hat{f}_K$ to approximate $f_K$ by combining data from  $K$ fidelities, each with $N$ samples (input parameters)  $\{\x_{k,n},y_{k,n}\}_{k=1,n=1}^{K,N}$.

\paragraph{Neural Processes.}

Combining Gaussian Processes (GPs) and neural networks (NNs), Neural Processes (NPs) \citep{garnelo2018neural} constitute a family of latent variable models for implicit stochastic processes \citep{wang2020doubly}. NPs represent distributions over functions and estimate prediction uncertainties like GPs while featuring scalability in high dimensions \citep{jha2022neural}.

Formally, NP consists of latent variables $z \in \mathbb{R}^{d_z}$ and model parameters $\theta$, trained on context set $\mathcal{D}^c \equiv \{x^c_{n},y^c_{n}\}_{n=1}^{N}$ and target sets $\mathcal{D}^t \equiv \{x^t_{m},{y^t_{m}}\}_{m=1}^{M}$. $\mathcal{D}^c$ and $\mathcal{D}^t$ are randomly split from the training set $\mathcal{D}$. Learning the posterior of $z$ and $\theta$ equals to maximize the posterior likelihood below:
%
\begin{equation}
p(y^t_{1:M}|x^t_{1:M},\mathcal{D}^c,\theta) = \int p(z|\mathcal{D}^c,\theta)\prod^{M}_{m=1}p(y^t_{m}|z, x^t_{m},\theta)dz 
\label{eqn:np}
\end{equation}
%
%
Due to the intractability of marginalizing over latent variables $z$, the NP family  utilize approximate inference, deriving the approximated evidence lower bound (ELBO):

\vskip -0.12in
\begin{align}
    & \log p(y^t_{1:M}|x^t_{1:M},\mathcal{D}^c,\theta) \gtrapprox \nonumber\\ 
    & \mathbb{E}_{q_\phi(z|\mathcal{D}^c \cup \mathcal{D}^t)} \big[  \sum_{m=1}^M\log p(y^t_m|z, x^t_m,\theta)+\log\frac{q_\phi(z|\mathcal{D}^c)}{q_\phi(z|\mathcal{D}^c \cup \mathcal{D}^t)}\big] 
\end{align}
This variational approach approximates the intractable true posterior $p(z|\mathcal{D}^c, \theta)$ with the approximate posterior $q_\phi(z|\mathcal{D}^c)$. Here, $\phi$ parameterize the encoder and $\theta$ parameterize the decoder of the model. Implementation-wise, each pair $\{x_n^c,y_n^c\} \in \mathcal{D}^c$ is first encoded as a latent representation $r^c_n$, forming a latent representation set $\{r_n^c\}^{N}_{n=1}$, then aggregated to parameterise the latent variable $z$. For simplicity, we denote $\{x_n\}^N_{n=1}$ as $X$ and $\{y_n\}^N_{n=1}$ as $Y$.

\section{Methodology}
\begin{figure*}[t!]
    \centering
    \includegraphics[width=1\linewidth,trim={0, 0, 0, 0}]{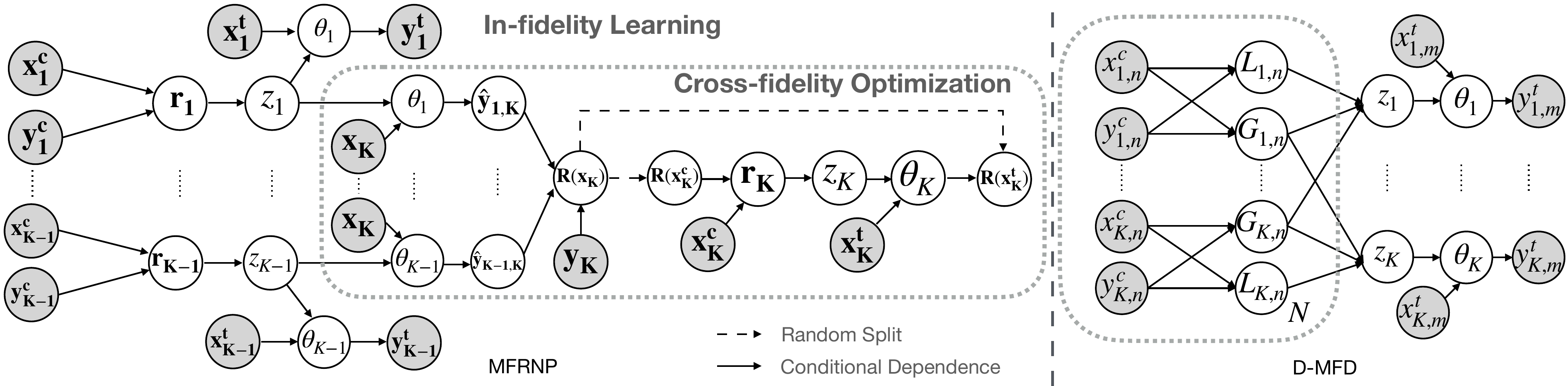}
    \caption{ Graphical model comparison of \ours{} against the state-of-the-art D-MFD baseline model. Shaded circles denote observed variables. Right: D-MFD disentangles the latent representations $r_{k,n}$ into local and global representations $L_{k,n}$ and $G_{k,n}$ to infer $z_k$. Each $G_{k,n}$ is from a different fidelity level, while no information about the decoder parameters $\theta_k$ is shared, making $G_{k,n}$ inaccurate regarding $\theta_i$ where $i\neq k$. Left: \ours{} dynamically constructs $\mathcal{D}^{train}_K = \{X_K, R(X_K)\}$ in cross-fidelity optimization step and learn fidelity-specific information with in-fidelity learning step. The residual function $R$ makes $z_K$ dependant on $z_{1:K-1}$ and $\theta_{1:K-1}$ without inflating any dimensions. Thus, \ours{} optimizes lower fidelity decoders for better information sharing. We use bold letters in our model graph to denote a set of variables.}
    \label{fig:struc}
\end{figure*}

The high-level goal of \ours{} is to explicitly model the residual between the aggregated output from the lower fidelities and the ground truth in the highest fidelity. It optimizes the aggregation to simultaneously promote in-fidelity learning and accurate cross-fidelity information sharing.

For each fidelity $k \in \{1...K-1\}$, \ours{} learns a NP surrogate $\hat{f}_{k}$ to approximate $f_{k}$, producing an output $\hat{y}_k$. For the highest fidelity $K$, \ours{} learns a NP to approximate the residual function 
$R(x) = f_K(x) - [\frac{1}{K-1}\sum_{k=1}^{K-1}\hat{f}_{k}(x)]$. 
The lower fidelity predictions are aggregated by linearly interpolating to the resolution at the highest fidelity and taking the average. We choose to average over other aggregation methods because this stabilizes the aggregation and ensures the equal contribution of lower fidelity surrogates. We explore other aggregation techniques in section \ref{sec:abl}. Unlike previous work where all the information from a fidelity is shared via the latent variable during encoding, \ours{} shares information via the decoded outputs. This explicitly involves the decoder at each fidelity to improve information-sharing. The input-specific information-sharing scheme further tackles the error propagation issue and ensures the accuracy and full exploitation of lower-fidelity information.

\textbf{Approximating the Residual Function.}
The core idea of \ours{} is to make the highest fidelity latent variable $z_K$ dependent on all other latent variables as well as the decoder parameters at lower fidelities. We introduce the dependency by approximating $R(x)$ at the highest fidelity, which models the residual between ground truth and the proposed aggregation from lower fidelities given input $x$. 

Intuitively, $\hat{f}_{k\in \{1,...,K-1\}}$ is optimized with two objectives. As shown in Figure \ref{fig:struc}, the in-fidelity learning step captures fidelity-specific information while the cross-fidelity optimization step encourages lower fidelity surrogates to propose accurate and informative aggregations w.r.t the highest fidelity given input $x_K$.

During inference,  given target input $x^t$, \ours{} approximates $f_K(x^t)$ by aggregating predictions from lower fidelities ($1...K-1$) and adding it with the residual term from fidelity $K: \hat{f}_{K}(x^t) = R(x^t) + [\frac{1}{K-1}\sum_{k=1}^{K-1}\hat{f}_{k}(x^t)]$. \ours{} generates $\hat{y}_K^t$ by fully exploiting the rich information in $\{\hat{y}_{1 \cdots K-1}\}$. In the OOD setup where $\mathcal{D}^{train}_K$ and $\mathcal{D}^{test}_K$ covers different input domains, \ours{} can effectively explore the regions beyond the $\mathcal{D}^{train}_K$, thus enhancing its input-domain extrapolation capabilities in modeling $f_K$.

\textbf{Residual-ELBO.} We design a Residual-ELBO (R-ELBO) for \ours{}. For each fidelity $k \in \{1 \cdots K-1\}$, we infer the latent variable $z_k$ with the NN encoder $q_{\phi_k}(z_k|\mathcal{D}^c)$ and decoder $p_{\theta_k}(y^t_{k}|z_k,x^t_k)$. At fidelity $K$, 
$\mathcal{D}_K$ is dynamically constructed with $\{X_K, R(X_K)\}$. Here $X_K$ denotes a set of input variables $x_K$. We infer the latent variable $z_K$ with the NN encoder $q_{\phi_K}(z_K|z_{1..K-1}, \theta_{1..K-1}, \mathcal{D}_{K}^c)$ and decoder $p_{\theta_k}(R(X_K)|z_{K}, X_{K},\theta_K)$,
where $R(X_K)$ depends on the proposed aggregations of lower fidelity predictions via ancestral sampling \cite{wang2020doubly}. 
We derive the R-ELBO for $K>1$ fidelities in two terms:
\vskip -0.12in

{\allowdisplaybreaks
\begin{align}
    & \small\log p(R(x_K^t)|x^t_{K},\mathcal{D}_{1: K}^c \cup \mathcal{D}_{1: K-1}^t,\theta_K) \nonumber\\
\geq&  \small\mathbb{E}_{q_{\phi_K}(z_{K}|\mathcal{D}_{1: K}^c \cup \mathcal{D}_{1: K}^t)} \nonumber\\
& \small\big[\log p(R(x_K^t)|z_{K}, x^t_{K},\theta_K)   \nonumber\\
    & \small+ \log \frac{q_{\phi_K}(z_{K}| \mathcal{D}_{1: K}^c \cup \mathcal{D}_{1: K-1}^t)}
    {q_{\phi_K}(z_{K}|\mathcal{D}_{1: K}^c \cup \mathcal{D}_{1: K}^t)} \big]\nonumber\\
    =& \small\mathbb{E}_{q_{\phi_K}(z_K|z_{1:K-1}, \theta_{1:K-1}, \mathcal{D}_{K}^c \cup \mathcal{D}_{K}^t))} \nonumber\\
    & \small\big[ \log p(R(x_K^t)|z_{K}, x^t_{K},\theta_K) \nonumber\\
    & \small+ \log \frac{q_{\phi_K}(z_K|z_{1:K-1}, \theta_{1:K-1}, \mathcal{D}_{K}^c)}{q_{\phi_K}(z_K|z_{1:K-1}, \theta_{1:K-1}, \mathcal{D}_{K}^c \cup \mathcal{D}_{K}^t))} \big] 
    \label{eqn:elbo_residual}
\end{align}
}
\vspace{-1.5\baselineskip}

\begin{align}
    & \small\log p(y^t_{1:K-1}|x^t_{1: K-1},\mathcal{D}_{1: K-1}^c,\theta_{1:K-1}) \nonumber\\
\geq&  \small\mathbb{E}_{q_\phi(z_{1: K-1}|\mathcal{D}_{1: K-1}^c \cup \mathcal{D}_{1: K-1}^t)} \nonumber\\
& \small\big[\log p(y^t_{1: K-1}|z_{1: K-1}, x^t_{1: K-1},\theta_{1:K-1})   \nonumber\\
    & \small+..+ \log \frac{q_\phi(z_{1: K-1}| \mathcal{D}_{1: K-1}^c)}
    {q_\phi(z_{1: K-1}|\mathcal{D}_{1: K-1}^c \cup \mathcal{D}_{1: K-1}^t)} \big]\nonumber\\
    =& \small\mathbb{E}_{q_{\phi_1}(z_1|\mathcal{D}_1^c \cup \mathcal{D}_1^t)..q_{\phi_{K-1}}(z_{K-1}|\mathcal{D}_{K-1}^c \cup \mathcal{D}_{K-1}^t)}\nonumber\\
    & \small\big[ \log p(y^t_1|z_1, x_1^t,\theta_1) \nonumber\\
    & +..+\log p(y_{K-1}^t|z_{K-1}, x_{K-1}^t,\theta_{K-1}) \nonumber\\
    & \small+ \log \frac{q_{\phi_1}(z_1|\mathcal{D}_1^c)}{q_{\phi_1}(z_1|\mathcal{D}_1^c \cup \mathcal{D}_1^t)}  \nonumber\\
    & \small+.. + \frac{q_{\phi_{K-1}}(z_{K-1}|\mathcal{D}_{K-1}^c)}{q_{\phi_{K-1}}(z_{K-1}|\mathcal{D}_{K-1}^c \cup \mathcal{D}_{K-1}^t)} \big] 
    \label{eqn:elbo}
\end{align}

Equation \ref{eqn:elbo} is a unified ELBO accounting for learning from fidelity specific datasets at lower fidelities. Equation \ref{eqn:elbo_residual} is the ELBO for the residual function at the highest fidelity. This term introduces dependency to every fidelity and optimizes the output aggregations to better approximate $R(x)$. 

For training, we calculate the R-ELBO with Monte Carlo (MC) sampling and ancestral sampling (AS) to optimize the objective function below: 
\begin{eqnarray}
    \mathcal{L}^{\hat{f}}_{MC} & = \sum^{K-1}_{k=1}\bigg[\frac{1}{S}\sum^{S}_{s=1}\log p(y_k^t|x_k^t,z_k^{(s)})
    \nonumber\\
    & - \text{KL}[q(z_k|\mathcal{D}_k^c, \mathcal{D}_k^t)\|p(z_k|\mathcal{D}_k^c)] \bigg]
    \label{eqn:L_i}
\end{eqnarray}
\begin{eqnarray}
    \mathcal{L}^R_{MC} & = \frac{1}{S}\sum^{S}_{s=1}\log p(R(x_K^t)|x_K^t,z_K^{(s)})
    \nonumber\\
    & - \text{KL}[q(z_K|\mathcal{D}_K^c, \mathcal{D}_K^t)\|p(z_K|\mathcal{D}_K^c)]\\\nonumber
    \label{eqn:L_ii}
\end{eqnarray} 
\begin{eqnarray}
    \mathcal{L}_{MC} & = \mathcal{L}^R_{MC} + \mathcal{L}^{\hat{f}}_{MC}
\end{eqnarray}
where the time for $q_{\phi}(z|\mathcal{D}^c)$ to sample $z_k^{(s)}$ scales linearly with the number of fidelity levels.

\textbf{Training \& Inference.}
As shown in Algorithm \ref{algo:MFRNP_train}, \ours{} calculates $\mathcal{L}^{\hat{f}}_{MC}$ by predicting the set $\hat{Y}_k^t$ given input target set $X_k^t$ for every fidelity $k \in \{1..K-1\}$. For $\mathcal{L}^R_{MC}$, the AS steps are highlighted in orange in Algorithm \ref{algo:MFRNP_train}.

\ours{} introduce decoder to the information aggregation step by predicting with the highest fidelity level input set $X_K$ at lower fidelity levels. We note the prediction as set $\{\hat{o}_{k,K}\}$ where $k\in\{1..K-1\}$, representing surrogate predictions given $X_K$ at fidelity $k$. We linearly interpolate each $\{\hat{o}_{k,K}\}$ to math the dimension of $Y_K$. 
Then we dynamically construct the training dataset at fidelity $K$ as $\mathcal{D'}_K = \{X_K,R(X_K)\}$. Finally, we randomly split $\mathcal{D'}_K$ into $\mathcal{D'}_K^c, \mathcal{D'}_K^t$ to obtain $\mathcal{L}^R_{MC}$ and perform back propagation with $\mathcal{L}_{MC}$. Here, $\mathcal{L}^R_{MC}$ encourages \ours{} to optimize lower fidelity surrogates for residual modeling at the highest fidelity,
while $\mathcal{L}^{\hat{f}}_{MC}$ regulates lower fidelity levels to learn from fidelity-specific datasets. 

The inference process is demonstrated in Algorithm \ref{algo:MFRNP_inference} where the AS steps are highlighted in orange. Given input set $\{x_{i,K}\}$, \ours{} propose aggregated predictions $\{a_{i,K}\}$ from lower fidelities and the predicted residual $\hat{R}_{i}$ at fidelity $K$. The final prediction is given as $\{\hat{y}_{i,K}\} = \{a_{i,K}\} + \{\hat{R}_{i}\}$.

\begin{algorithm}[t!]
\begin{algorithmic}
\STATE {\bfseries Input:} Dataset $\mathcal{D}_{1\cdots K}$, number of fidelities $K$.
\FOR{$k=1$ {\bfseries to} $K-1$}
 \STATE Randomly split $\mathcal{D}_{k}$ into $\{\mathcal{D}_{k}^c, \mathcal{D}_{k}^t\}$
 \STATE    Sample $\{z_{i,k}\}$ where $ z_{i,k} \sim q_{k}(.|\mathcal{D}_k^c)$\\
 \STATE    Predict $\{\hat{y}_{i,k}^t\}$ where $ \hat{y}_{i,k}^t \sim p_{k}(.|z_{i,k}, x^t_{i,k})$\\
 \STATE    \textcolor{orange}{ 
           Sample $\{z'_{i,k}\} $ where $ z'_{i,k} \sim q_{k}(.|\mathcal{D}_k^c)$\\
\STATE    Predict $\{\hat{o}_{i, k}\} $ where $\hat{o}_{i, k} \sim 
p_{k}(.|z'_{i,k}, x_{i,K})$\\
 \STATE    Linearly interpolate $\{\hat{o}_{i, k}\}$ to the resolution at $K$.}\\
\ENDFOR
\STATE Get the residual set $\{R_i\}$ at K where $R_i = y_{i,K} - \frac{\sum(\hat{o}_{i,1},..,\hat{o}_{i,K-1})}{K-1}$\\
\STATE Random-split $\mathcal{D'}_K=\{x_{i,K},R_{i}\}$ into $\{\mathcal{D'}_K^c, \mathcal{D'}_K^t\}$\\
\STATE    Sample $\{z_{i,K}\}$ where $z_{i,K} \sim q_{K}(.|\mathcal{D}_K^c)$\\
\STATE    Predict $\{\hat{R}^t_{i}\}$ where $\hat{R}^t_{i} \sim p_{K}(.|z^t_{i,K}, x^t_{i,K})$\\
\STATE Back propagate with $\mathcal{L}_{MC} = \mathcal{L}^R_{MC} + \mathcal{L}^{\hat{f}}_{MC}$ in Eqn \ref{eqn:L_ii}\\
\end{algorithmic}
 \caption{MFRNP Training Process}
 \label{algo:MFRNP_train}

\end{algorithm}

\begin{algorithm}[t!]
\begin{algorithmic}
\STATE {\bfseries Input:} Latent variables $z_{1..K}$, input $\{x_{i,K}\}$ 
\FOR{$k=1$ {\bfseries to} $K-1$}
 \STATE \textcolor{orange}{ 
           Sample $\{z'_{i,k}\}$ from $z_k$.\\
  \STATE   Predict $\{\hat{o}_{i, k}\} $ where $\hat{o}_{i, k} \sim 
p_{k}(.|z'_{i,k}, x_{i,K})$\\
  \STATE   Linearly interpolate $\{\hat{o}_{i, k}\}$ to the resolution at $K$.}\\
\ENDFOR
\STATE Obtain aggregation $\{a_{i,K}\} = \frac{\sum(\hat{o}_{i,1},..,\hat{o}_{i,K-1})}{K-1}$\\
\STATE   Sample $\{z'_{i,K}\}$ from $z_K$.\\
\STATE Predict $\{\hat{R}_{i}\}$ where $\hat{R}_{i} \sim p_{K}(.|z_{i,K}, x_{i,K})$\\
\STATE   Return $\{\hat{y}_{i,K}\} = \{a_{i,K}\} + \{\hat{R}_{i}\}$
\end{algorithmic}
 \caption{MFRNP Inference Process}
 \label{algo:MFRNP_inference}

\end{algorithm}
\section{Related Work}
\paragraph{Multi-fidelity Modeling.}
Multi-fidelity surrogate modeling is prevalent across scientific and engineering domains, including applications in aerospace systems \cite{brevault2020overview} and climate science \cite{Hosking2020,valero2021multifidelity}. The foundational work of \citet{kennedy2000predicting} employs GPs to connect models of varying fidelity levels, introducing an autoregressive model. \citet{le2014recursive} introduces a recursive GP with a nested structure in the input domain to facilitate rapid inference. \citet{perdikaris2015multi, perdikaris2016multifidelity} addresses high-dimensional GP scenarios by leveraging the Fourier transformation of the kernel function. \citet{perdikaris2017nonlinear} puts forth the concept of multi-fidelity Gaussian processes (NARGP), assuming a nested structure in the input domain for sequential training at each fidelity level.

\citet{wang2021multi} presents a Multi-Fidelity High-Order GP model for accelerating physical simulations. They extend the Linear Model of Coregionalization (LMC) to the nonlinear case, incorporating a matrix GP prior on the weight functions. Deep Gaussian processes (DGPs) \citep{cutajar2019deep} formulate a unified objective to optimize kernel parameters jointly across fidelity levels. DGPs face scalability challenges with high-dimensional data. Infinite-Fidelity Coregionalization (IFC) from \cite{li2022infinite} models the output space as a continuous function of fidelity and input based on neural ODEs, allowing the model to extrapolate to higher fidelities. However, IFC faces scalability issue due to the computation-demanding ODE solver.

Multi-fidelity modeling has been integrated with deep learning.
\citet{guo2022multi} employs deep neural networks to merge parameter-dependent output quantities. \citet{wang2023diffusion} utilizes diffusion model to fuse fidelity information into the diffusion-denoising process to solve PDE problems with a conditional scoring model. \citet{meng2020composite} proposes a composite neural network for multi-fidelity data in inverse PDE problems, while \citet{meng2021multi} introduces Bayesian neural nets for multi-fidelity modeling. \citet{de2020transfer} employs transfer learning to fine-tune high-fidelity surrogate models using deep neural networks trained on low-fidelity data. Despite advancements, existing deep GP models \citep{cutajar2019deep,hebbal2021multi} struggle with cases where different fidelities involve data with varying dimensions. Additionally, multi-fidelity methods have been investigated in Bayesian optimization, active learning, and bandit problems \citep{li2020multi, song2019general, li2022batch,perry2019allocation,kandasamy2017multi}.

\paragraph{Neural Processes.}
Neural Processes (NPs) \cite{garnelo2018conditional, kim2018attentive, louizos2019functional,singh2019sequential} emerge as scalable and expressive alternatives to GPs for modeling stochastic processes. Previous work by \citet{raissi2016deep} combines multi-fidelity GPs with deep learning, placing a GP prior on features learned by deep neural networks. \citet{wang2020mfpc} proposes a multi-fidelity neural process with physics constraints (MFPC-Net), leveraging NPs to capture correlations between multi-fidelity data. Nonetheless, this model requires paired data and cannot utilize unpaired data at the low-fidelity level.

The recent work of \citet{wu2023disentangled} proposes to disentangle the latent variable at each fidelity into local and global representations and share the global part across fidelities. However, the fidelity-specific decoder parameters  are not included in information sharing. Thus, the highest fidelity decoder expresses shared representations differently, hindering the inference performance, especially in OOD scenarios.

\paragraph{Climate Modeling.} \label{relate}
Climate modeling is a central component of modern climate science and the primary tool for predicting future climate states \cite{flato2014evaluation}. Various modeling centers around the world have developed distinct climate models. The semi-independent development process has led to many plausible, but disagreeing, climate models representing the same earth system \cite{knutti2010challenges,flato2014evaluation}. Averaging these models often leads to improved results compared to using individual models \cite{lambert2001cmip1, gleckler2008performance, knutti2010challenges}. Proper aggregation of different climate models for a consensus estimate is therefore an important topic \cite{tebaldi2007use}. Averaging all models with equal or varying weights has been the most common approach \cite{giorgi2002calculation, abramowitz2019esd}, known as (weighted) ensemble averaging (EA). However, EA techniques often do not retain much spatial information and can cause severe blurring, corrupting regional signals. 

To address this problem, various alternative approaches have been developed based on Bayesian hierarchical models, regression and machine learning that all use observational data to improve model aggregation. DeepESD \cite{bano2022downscaling} utilize CNNs to learn the mapping from the EA of $8$ regional climate models to the observation-calibrated data \cite{dee2011era} at finer resolution. Similarly, NNGPR \cite{harris2023multimodel} utilizes deep kernel GPs to model the residual between the EA of $16$ low-resolution global climate models and the up-to-date observation-calibrated ERA5-reanalysis dataset \cite{hersbach2020era5}. A lot of progress have been made in downscaling the climate simulators with observational data \cite{kotamarthi2021downscaling}, yet lacking the method to accurately infer the long-term climate scenarios directly from climate drivers.

\section{Experiments}

\begin{table*}[h]
\small
\centering
\begin{tabular}{l|l|l|l|l|l|l}
\toprule
Task (Full) & DMF & NARGP & MFHNP & D-MFD & SF-NP & \ours{}\\ \midrule
{Heat 2}    & 0.138 ± 4.0e-8 & 0.31±2.12e-6    & 0.026±4.01e-5 & 0.015±1.42e-5 & 0.308±6.38e-05 &\textbf{0.005±3.27e-4}   \\ 
\midrule
{Heat 3}       & 0.137±1.23e-7     & 0.309±3.46e-6    & 0.111±4.82e-6 & 0.108±4.85e-8 &  0.307±8.62e-05 & \textbf{0.0039±2.94e-4}   \\ 
\midrule

{Heat 5}       &   0.135±2.55e-4   & 0.306±1.14e-4   & 0.115±1.22e-2  & 0.106±1e-4
 & 0.306±6.90e-05 & \textbf{0.0045±2.94e-4}   \\ 
\midrule

{Poisson 2}       & 0.107 ± 6.58e-5     & 0.585±9.84e-5    & 0.093±2.55e-4 & 0.07 ±2.99e-4 & 0.575±1.39e-04 & \textbf{0.0076±7.49e-4}  \\ 
\midrule
{Poisson 3}       & 0.121±1.47e-5     & 0.58 ±1.02e-4    & 0.335±2.37e-5 & 0.101±1.81e-4 & 0.572±1.95e-04 &  \textbf{0.0073±5.25e-4} \\ 
\midrule

{Poisson 5}       &  0.101±2.24e-3    & 0.571±1.29e-4    & 0.299±8.84e-3 & 0.279±3.35e-3 & 0.571±1.82e-04  &  \textbf{0.0046±1.2e-4} \\ 
\midrule
{Fluid}       & 0.275±4.59e-7     & 0.353±9.28e-4    & 0.234±4.82e-6 & 0.207±1.31e-5 &  0.383±5.53e-05  & \textbf{0.129±8.19e-4}  \\ \bottomrule
\end{tabular}
\caption{Performance (nRMSE) comparison of 6 different models applied to the Heat and Poisson simulators with two, three, five fidelities and fluid simulation with Navier-Stokes equation with two fidelities. The full setting means same domain coverage of $\mathcal{D}_K^{train}$ and $\mathcal{D}_K^{test}$.}
\label{tb:full}
\end{table*}
\begin{table*}[h]
\small
\centering
\begin{tabular}{l|l|l|l|l|l|l}
\toprule
Task (OOD) &  DMF & NARGP & MFHNP & D-MFD & SF-NP & \ours{}\\ \midrule
{Heat 2}       &  0.168±1.36e-4     & 0.313±1.30e-4    & 0.033±1.37e-2 & 0.213±1.65e-3 & 0.312±1.14e-04 & \textbf{0.005±1.33e-4}   \\ 
\midrule
{Heat 3}       &  0.163±6.04e-4   & 0.309±5.00e-4   & 0.143±5.88e-3 & 0.141±4.94e-3 & 
0.310±7.03e-05 &\textbf{0.004±3.82e-4}   \\ 
\midrule
{Heat 5}       &  0.187±8.60e-4   &  0.308±1.05e-4  & 0.15±2.72e-3  & 0.145±2.87e-3 & 0.308±5.25e-05 & \textbf{0.012±1.05e-2}\\ 
\midrule
{Poisson 2}       &   0.183±6.85e-4   &  0.749±1.47e-5  & 0.103±2.03e-2  & 0.214±4.28e-2  & 0.749±1.79e-03 &\textbf{0.017±2.72e-3}  \\ 
\midrule
{Poisson 3}       &   0.186±8.06e-4   & 0.744±8.24e-5    & 0.189±1.14e-2 & 0.2±1e-2 & 0.745±7.62e-04 &  \textbf{0.018±1.27e-3} \\ 
\midrule

{Poisson 5}       &  0.16±4.47e-4    &  0.743±3.07e-4  & 0.399±1.07e-2   &  0.375±8.56e-3 & 0.744±1.85e-03 &   \textbf{0.013±2.76e-4}\\ 
\midrule

\end{tabular}
\caption{Performance (nRMSE) comparison of 6 different models applied to the Heat and Poisson simulators with two, three, five fidelities. The OOD setting here indicates $\mathcal{D}_K^{test}$ is OOD w.r.t the training domain at fidelity $K$.}

\label{tb:OOD}
\end{table*}

\subsection{Datasets} 
We include 6 Partial Differential Equation benchmarks, a more complicated fluid simulation task and a real-world climate modeling task for earth surface temperature prediction.
\paragraph{\textbf{Partial Differential Equations (PDEs).}} We include Heat and Poisson's equations \cite{olsen2011numerical} from computational physics. We test \ours{} for predicting the spatial solution fields of these equations. We use numerical solvers to generate the ground-truth data with dense and coarse meshes for different fidelity levels. For both Heat and Poisson's equations with $2$ fidelity setting, we use $16 \times 16$ and $32 \times 32$ meshes as low and high fidelities. For three fidelity scenarios, we run the solvers with $64 \times 64$ meshes at the highest fidelity level. For five fidelity scenarios, we run the solvers with $96 \times 96$ and $128 \times 128$ meshes as the two additional fidelity levels. The heat equation has an input dimension of 3, corresponding to the thermal diffusivity coefficient and boundary conditions at the two edges. The Poisson's equation has an input dimension of 5, corresponding to the $4$ boundary conditions and the magnitude of flow at the centered point source.

\paragraph{\textbf{Fluid Simulation.}}

This task simulates the dynamics of a circular smoke cloud propelled by an inflow force within a $50 \times 50$ grid. The spatial-temporal solution field is derived through the application of the incompressible Navier-Stokes equations and the Boussinesq approximation \cite{holl2020learning, chorin1968numerical}. Initiated with the introduction of an incompressible static smoke cloud of radius $5$ at the lower center of the grid, a persistent inflow force is subsequently applied at the center of the cloud. The input parameters encompass two variables governing the magnitude of the inflow force along the $x$ and $y$ directions. The output is the initial component of the velocity field following a temporal evolution of $30$ steps with the inflow force. The simulation generates low-fidelity ground-truth with $32 \times 32$ mesh, and high-fidelity with $64 \times 64$ mesh.

\paragraph{\textbf{Climate Modeling: Earth Surface Temperature.}}
In this task, we take one step further from previous works in Section \ref{relate} with multi-fidelity surrogate modeling to directly learn the mapping from climate drivers \cite{watson2022climatebench} to the observation-calibrated ERA5-reanalysis data \cite{hersbach2020era5}, together with $13$ low-resolution computational climate model predictions from the ScenarioMIP project\cite{o2016scenario} at global scale. We provide details about the computational models used in Appendix \ref{climate_ds}.

We group these climate data into $9$ fidelities based on their original resolutions. The fidelity dimensions are: $144 \times 192$, $160 \times 320$, $192 \times 288$, $180 \times 288$, $120 \times 180$, $143 \times 144$, $80 \times 96$, $192 \times 384$ and $721 \times 1440$ at the highest fidelity level, representing air temperature at 2 meters above the earth surface. The climate drivers as inputs here consists of $12$ variables representing the total emission of greenhouse gases (CO2, CH4) and aerosol gases (BC, SO2) from year $1850$ to $2015$. Each of the aerosol gas is split into $5$ signals via principal component analysis \cite{wold1987principal, watson2022climatebench}. For years after $2015$, we test $4$ hypothetical global gas emission scenarios: ssp126, ssp245, ssp370, ssp585. Larger numbers correspond to more total gas emissions, leading to more severe climate change. Ssp126 represents the condition in which future gas emissions are well controlled and gradually decrease. This scenario makes the gas emission data as input to have similar domain coverage as of training data. Whereas in the other three scenarios, future gas emissions would keep increasing, introducing out-of-distribution inputs at the highest fidelity.

\begin{figure*}[t!]
    \centering
    \includegraphics[width=1\linewidth,trim={0, 0mm, 0, 0}]{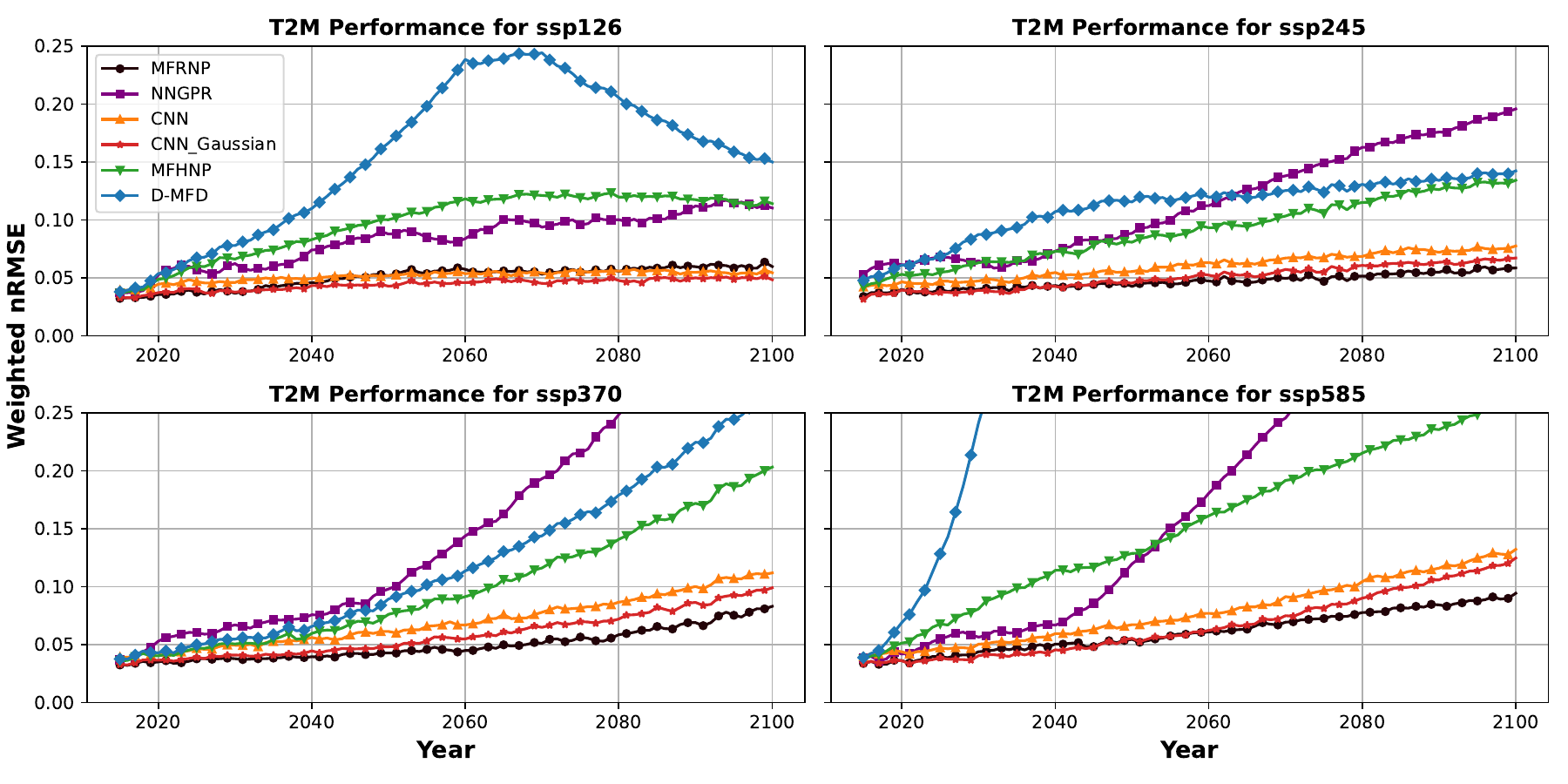}
    \vspace{-3mm}
    \caption{Perfect model test performance for climate modeling task. Measured in latitude-weighted nRMSE. \ours{} outperforms other models in 3 out of the 4 scenarios and maintains consistent performance across all the scenarios from year $2015\sim2100$.
    }
    \label{fig:pm_test}
\end{figure*}

\subsection{Experiment Setup}
We implement \ours{} with PyTorch \cite{paszke2019pytorch} and compare the average nRMSE in three random runs with following baselines. Details are provided in Appendix \ref{Exp_Setup}. 
\begin{itemize}
  \itemsep0.2em
  \item \textbf{SF-NP.} The naive single-fidelity neural process trained on the highest fidelity data, as the lower performance bound for multi-fidelity modeling.
  \item \textbf{DMF}\cite{li2021deep} performs multi-fidelity learning based on a multi neural network (NN) structure. Each NN corresponds to one fidelity and the NN on the next fidelity adapts the latent output from the NN at current fidelity to propagate information. 
  \item \textbf{NARGP}\cite{perdikaris2017nonlinear} utilizes multi-fidelity Gaussian Processes by sequentially train Gaussian Processes from low to high fidelity levels under the assumption of a nested input domain.
  \item \textbf{MFHNP}\cite{wu2022multi} learns distribution over functions at each fidelity level with a latent variable. They designed a hierarchical structure of the latent variables from low to high fidelity levels to pass information.
  \item \textbf{D-MFD}\cite{wu2023disentangled} tackles the error propagation problem in MFHNP and the inter-fidelity NN overfitting problem in DMF by introducing local and global latent representations at each fidelity. \\
\end{itemize}

\paragraph{PDEs Setup.} 
For PDE tasks, we consider two data composition scenarios.
\begin{itemize}
    \itemsep0.2em
    \item \textbf{In-distribution (Full) Scenario.} In this setting, the training set $\mathcal{D}^{train}$ and testing set $\mathcal{D}^{test}$ at each fidelity level have the same coverage on dataset domain $\mathcal{D}$.

    \item \textbf{Out-of-distribution (OOD) Scenario.} This scenario simulates real-world conditions where the highest fidelity data has limited domain coverage. We evaluate model trained using $\mathcal{D}^{train} \subset \mathcal{D}$ at the highest fidelity level, and test the model performance on $\mathcal{D}^{test}$ consists of data covering the rest of $\mathcal{D}$ s.t. $\mathcal{D}^{train} \cap \mathcal{D}^{test} = \emptyset$ and $\mathcal{D}^{train} \cup \mathcal{D}^{test} = \mathcal{D}$.

\end{itemize}
We construct the OOD scenario with Heat and Poisson's equations. We did not include Fluid data because the low fidelity mesh is not detailed enough to capture the differences between $\mathcal{D}^{train}$ and $\mathcal{D}^{test}$. Under the OOD setting, information covering the input domain of $\mathcal{D}^{test}$ is missing in all the fidelities, making it impossible for models to extrapolate with lower fidelity information.
For Heat equation, we test the OOD scenario of the thermal diffusivity coefficient by limiting the corresponding input parameter to cover $80\%$ of original domain. We uniformly sample $X$ in the constrained domain and obtain corresponding $Y$ with numerical solver as $\mathcal{D}^{train}$. $\mathcal{D}^{test}$ is constructed with the rest $20\%$ of original domain. Similarly, for Poisson's equation, we construct $\mathcal{D}^{train}$ and $\mathcal{D}^{test}$ with the $80/20$ coverage split on $3$ boundary conditions. Details are demonstrated in Appendix \ref{OOD_Data}.

\begin{figure*}[t!]
    \centering
    \includegraphics[width=1\linewidth]{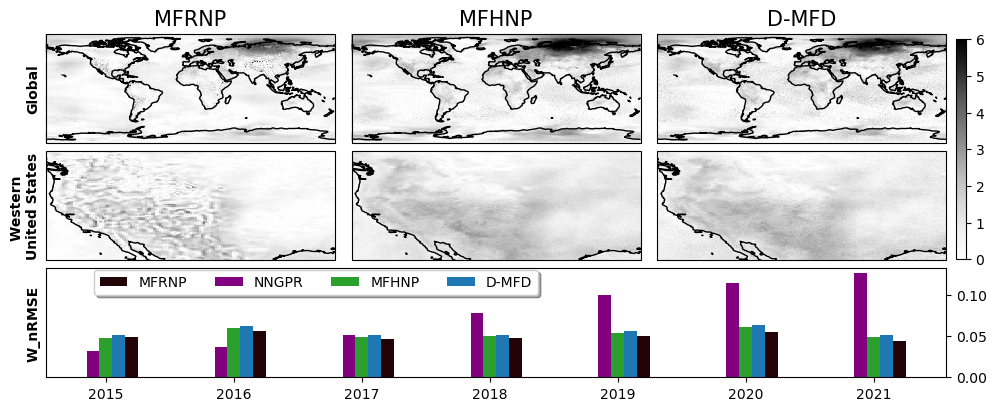}
    \vspace{-3mm}
    \caption{Absolute residual plot (°C) between ground truth and model predictions on ERA5-reanalysis data at year $2020$, with global resolution of $721\times1440$. The first row presents global view, second row focuses on smaller geographic regions. The third row shows the latitude weighted nRMSE to measure prediction accuracy from year 2015 to 2021. \ours{} outperforms other models staring from $2017$. 
    }
    \label{fig:era5_vis.pdf}
\end{figure*}

\paragraph{Climate Modeling Setup.} 
In this task, we also compare \ours{} with recent works in the climate science community of downscaling global climate projections. We include deep kernel GPs from NNGPR \cite{harris2023multimodel} and Gaussian CNN from DeepSED \cite{bano2022downscaling}. Since no previous work simultaneously downscales while modeling the mapping between climate drivers and scenarios, to ensure fair comparison under the same task, we set up a naive extension for these methods. We first fit $13$ linear surrogate models with the climate drivers as input $X$, and the computational climate model predictions as target $Y$. Then we train the models in NNGPR and DeepSED to downscale the averaged predictions from linear surrogates given $X$. 

We use $13$ "perfect model" tests \cite{knutti2017climate} as the performance metric. In each test, one climate model is held out as the ground truth and we only use its historical data ($1950\sim2015$) for training at the highest fidelity. The remaining $12$ climate models are grouped at lower fidelity levels based on their resolutions. We compare the average latitude weighted nRMSE across $13$ tests. Thus, we are able to empirically measure the performance with physics based future projections. We then include the ERA5-reanalysis data as the highest fidelity level to measure the performance on a refined resolution. We use year $1950\sim2014$ as the highest fidelity training data and $2015\sim2021$ for testing.

\subsection{Results}

\paragraph{PDEs Performance.}
We test the performance of \ours{} and baselines across $7$ tasks under the full setting and $6$ tasks under the OOD setting. Table \ref{tb:full} shows the full setting performance and Table \ref{tb:OOD} shows the OOD setup performance. Among all the tasks and settings, our model consistently outperforms baselines by an average of $\sim90\%$ in nRMSE. 

The results indicate that \ours{} is more efficient in utilizing information across fidelities to generate good predictions. Unlike other baselines for which the performance generally gets worse as the number of fidelities increase, the performance of \ours{} in the $5$ fidelity tasks is even better than the corresponding $2$ fidelity tasks in $3$ out of $4$ five-fidelity experiments. This indicates \ours{} gathers lower fidelity information efficiently enough to outrun the increasing task complexity in modeling Heat and Poisson's equations.

\paragraph{Climate Modeling Performance.}
As shown in Figure \ref{fig:pm_test}, for the perfect model test, \ours{} performs consistently across all emission scenarios and outperforms other baselines in 3 out of the 4 scenarios. The perfect model test empirically shows the capability of \ours{} to efficiently aggregate information from lower fidelities and its versatility for OOD scenarios. Although CNN and Gaussian CNN perform relatively well, they have very limited scalability and cannot scale to the global ERA5-reanalysis dataset. We also had memory issues running NARGP and DMF with 9 fidelities for this task. Among the baseline models that scale to the ERA5-reanalysis dataset (MFHNP, D-MFD and NNGPR), the performance of \ours{} is significantly better.

For the ERA5-reanalysis data, \ours{} also outperforms other baseline models, as shown in Figure \ref{fig:era5_vis.pdf}. Our model keeps a consistent performance from year $2015\sim2021$, while the performance of other baseline models worsens for further future predictions. \ours{} makes a cleaner residual plot compared to the other models. 
Although we can only evaluate performance up to year $2021$ due to observation limitations, we have empirically shown the long term prediction consistency of \ours{} from the perfect model test and we expect it to further outperform other baselines for years after $2021$ with the ERA5-reanalysis data.

\ours{} has shown superior performance across various gas emission scenarios from year $2015$ to $2100$ via the perfect model test. Together with the accurate reanalysis predictions, we have demonstrated that \ours{} is not only capable of fully exploiting lower fidelity information for reliable predictions, but also able to provide great scalability with consistent performance for real-world tasks.

\subsection{Ablation Study \label{sec:abl}}

\paragraph{Latent Aggregation with Residual.} We evaluate the impact of including decoders in the information sharing step. The same experiment settings are maintained for consistency in the comparison. We create a variant of \ours{}, namely MFRNP-H, for which the surrogate models for fidelity levels below the highest are implemented with a hierarchical structure that shares information via the latent space \citep{wu2022multi}. We use the surrogate output at fidelity $K-1$ as the aggregated information and model the residual between this aggregation and the highest fidelity ground truth. 

\begin{table}[ht]
\centering
\begin{tabular}{l|l|l}
\toprule
Task (Full) & MFRNP-H & \ours{} \\ \midrule
{Heat 3}   &  0.013±4.46e-4  & \textbf{0.0039±2.94e-4} \\
\midrule
{Heat 5} & 0.013±2.76e-4 & \textbf{0.0045±2.94e-4}\\
\midrule
{Poisson 3}       & 0.138±6.1e-3  & \textbf{0.0073±5.25e-4}\\
\midrule

{Poisson 5} & 0.184±1.77e-2 &  \textbf{0.0046±1.2e-4}\\
\midrule
\end{tabular}
\caption{Performance (nRMSE) comparison of MFRNP-H and MFRNP applied to the Heat and Poisson simulators with three, five fidelities under Full setup.}
\label{tb:abl_struc}
\end{table}

\begin{table}[ht]
\centering
\begin{tabular}{l|l|l}
\toprule
Task (OOD) & MFRNP-H & \ours{}\\ \midrule
{Heat 3} &  0.047±2.72e-4  & \textbf{0.004±3.82e-4} \\
\midrule
{Heat 5} & 0.055±1.87e-4 & \textbf{0.012±1.05e-2}\\
\midrule
{Poisson 3} & 0.128±1.3e-2  & \textbf{0.018±1.27e-3} \\
\midrule
 {Poisson 5} & 0.151±7.9e-3 &  \textbf{0.013±2.76e-4}\\
\midrule
\end{tabular}
\caption{Performance (nRMSE) comparison of MFRNP-H and MFRNP applied to the Heat and Poisson simulators with three, five fidelities under OOD setup.}
\label{tb:abl_struc_OOD}
\end{table}

Our results in Table \ref{tb:abl_struc} and Table \ref{tb:abl_struc_OOD} show that \ours{} significantly outperforms MFRNP-H. This indicates including decoders in the information aggregation step promotes cross-fidelity information sharing and yields better learning capabilities for both Full and OOD setup.

\paragraph{Weighted Averaging.} We explore the impact of non-uniform averaging in this study on Heat and Poisson dataset with $5$ fidelities under Full setting. We tested \ours{} with uniform averaging for simplicity and to avoid introducing additional hyperparameters. Intuitively, higher fidelity levels should be assigned higher weights since they provide more detailed predictions. We set the weighted average scheme to correlate with fidelity level as following: 
\begin{equation}
\begin{split}
    \text{Aggregation} = & \, 0.1 \times \hat{Y}_{k=1} +  0.2 \times \hat{Y}_{k=2} \\
                         & +  0.3 \times \hat{Y}_{k=3} +  0.4 \times \hat{Y}_{k=4}
\end{split}
\end{equation}
\begin{table}[ht]
\vspace{-3mm}
\centering
\begin{tabular}{l|l|l}
\toprule
Task & Uniform Averaging & Weighted Averaging \\ 
\midrule
Heat5 & 0.0045±2.94e-4 & \textbf{0.0042±1.19e-3} \\
Poisson5 & \textbf{0.0046±1.2e-4} & 0.0103±2.02e-3 \\
\bottomrule
\end{tabular}
\caption{Comparison of Uniform Averaging and Weighted Averaging on Heat5 and Poisson5 tasks, measured with nRMSE.}
\label{tb:averaging_comparison}
\vspace{-3mm}
\end{table}

As shown in Table \ref{tb:averaging_comparison}, for Heat5, using increasing weights grants $6.67 \%$ performance improvement. However, for Poisson5, using increasing weights doubles the nRMSE. Fine tuning the fidelity weights could potentially bring performance improvements, but it would require dataset-specific weight adjustments.

\section{Discussion \& Conclusion}
In this paper, we present Multi-fidelity Residual Neural Processes (\ours{}), a novel Neural Process-based multi-fidelity surrogate model. \ours{} utilizes a residual modeling framework, which allows \ours{} to leverage the rich input space coverage from lower fidelities while preserving accuracy from the highest fidelity data. Our tailored Residual-ELBO loss promotes learning across fidelities and simultaneously optimizes the lower fidelity decoders for accurate information sharing. Experimental results on partial differential equations and climate modeling demonstrate that \ours{} outperforming state-of-the-art methods by more than $90\%$, highlighting scalability, efficiency in information fusion, and versatility of \ours{} for real-world surrogate modeling. 

The limitation of \ours{} mainly lies in encoder and decoder complexity. Due to ancestral samplings step, the inference time is doubled. Although this additional time penalty is small compared with the actual simulators, it is a potential concern for computationally expensive encoders and decoders, such as neural ODEs. Moreover, uniform averaging may cause loss of detailed information from lower fidelities. For future work, we plan to address this limitation and incorporate physics-informed structure to better model expensive physics simulations.

\section*{Acknowledgments}
This work was supported in part by the U.S. Army Research Office under Army-ECASE award W911NF-07-R-0003-03, the U.S. Department Of Energy, Office of Science, IARPA HAYSTAC Program, NSF Grants SCALE MoDL-2134209, CCF-2112665 (TILOS), \#2205093, \#2146343,  \#2134274,  CDC-RFA-FT-23-0069, as well as DARPA AIE FoundSci.

\newpage
\section*{Impact Statement}
The work in this paper aims to advance the performance and scalability of multi-fidelity surrogate modeling. The model proposed has potential impact in climate science, computational chemistry, structural engineering and many other communities, all of which, we foresee potential positive impacts in accelerating physics simulations and scientific discoveries.


\bibliography{ref}
\bibliographystyle{icml2024}



\newpage
\appendix
\onecolumn
\section{Appendix}
\subsection{Climate Modeling Task Dataset} \label{climate_ds}
We list the $13$ computational climate models used here, together with the literatures. They are all included in the ScenarioMIP project \cite{o2016scenario}. ACCESS-CM2\cite{bi2020configuration};
BCC-CSM2-MR\cite{wu2021bcc};
CMCC-CM2-SR5\cite{cherchi2019global};
GFDL-ESM4\cite{dunne2020gfdl};
INM-CM5-0\cite{volodin2022possible};
IPSL-CM6A-LR\cite{boucher2020presentation};
KACE-1-0-G\cite{byun2019nims};
MCM-UA-1-0\cite{stouffer2019ipcc};
MRI-ESM2-0\cite{yukimoto2019meteorological};
CESM2\cite{danabasoglu2020community};
MPI-ESM\cite{gutjahr2019max};
NorESM2\cite{seland2020overview};
UKESM1\cite{sellar2019ukesm1}.

\subsection{OOD Dataset Construction} \label{OOD_Data}
Generally, we follow the experiment setup as the full scenario. The following describes the difference in building the highest fidelity dataset $\mathcal{D}_K$.
For Heat equation, we uniformly sample $X_{train}$ in the training scope: ((0,0.8),(-1,0),(0.01,0.1)) and use numerical solver to generate the corresponding $Y_{train}$. We then uniformly sample $X_{test}$ in the testing scope ((0.8, 1),(-1,0),(0.01,0.1))
 and use numerical solver to generate $Y_{test}$.

For the Poisson's equation, we uniformly sample $X_{train}$ in the training scope:  ((0.1, 0.74),(0.1, 0.74),(0.1, 0.74),(0.1, 0.9),(0.1, 0.9))
and use numerical solver to generate the corresponding $Y_{train}$. We then uniformly sample $X_{test}$ in the testing scope ((0.74, 0.9),(0.74, 0.9),(0.74, 0.9),(0.1, 0.9),(0.1, 0.9))
 and use numerical solver to generate $Y_{test}$.

\subsection{Experimental Setup} \label{Exp_Setup}
\paragraph{Metrics.} We use latitude weighted nRMSE as shown in Equation \ref{w_nRMSE} to measure model performance in the climate modeling task. For PDE tasks, we use nRMSE as shown in Equation \ref{nRMSE}.

\begin{equation}
    W\_nRMSE = \frac{\sqrt{\sum_{i=1}^{N} \left(\frac{\cos({\text{lat}_i})}{\frac{\sum_{j=1}^{N}\cos({\text{lat}_j})}{N}}\right) * \left(\frac{(y_i - \hat{y}_i)^2}{N}\right)}}{std(\{y_{1:N}\})}
    \label{w_nRMSE}
\end{equation}

\begin{equation}
    nRMSE = \frac{\sqrt{\sum_{i=1}^{N} \left(\frac{(y_i - \hat{y}_i)^2}{N}\right)}}{std(\{y_{1:N}\})}
    \label{nRMSE}
\end{equation}

\paragraph{PDE task Training Configurations.} For training, we use Adam optimizer \cite{kingma2014adam} with base learning rate of $1e-3$. We use $10\%$ of training data set as validation set. For Heat and Poisson's equation, we run our model with latent dimension and encoder/decoder dimension of $32$, and run our model with maximum epoch of $50000$ and patience $10000$. We use learning rate decay of $0.85$ and stepsize $10000$. We set the highest fidelity weight to $2$ and lower fidelities to $1$ in loss calculation to focus more on optimizing toward the highest fidelity. For Poisson5, using lower fidelity weight of $0.25$ further improves performance. For context-target split, we randomly select $20\% \sim 25\%$ of training data as our context set, the rest as our target set for each fidelity. For fluid simulation, we follow the same setup but use latent dimension and encoder/decoder dimension of $128$. We set patience to $5000$ with learning rate decay of $0.01$. We normalize the data before training and measure nRMSE on the de-normalized space. For the baseline models, we follow the same setup as above. All models are trained on NVIDIA A100 GPU with 80GB memory.

\paragraph{Climate Modeling Training Configurations.} We follow the similar setup as above, but set the highest fidelity weight to $5$, latent and hidden dimensions to $512$ to incorporate more fidelities and data at higher dimensions. We do not normalize the data before training. For the climate methods introduced from climate science community (NNGPR, CNN, CNN\_Gaussian), we use their original model setup. The linear surrogates for low-resolution climate simulators are built with $4$ layers with ReLU activation. Dimensions are $12$, $\floor{\frac{(lat*lon)}{128}}$, $\floor{\frac{(lat*lon)}{32}}$, $\floor{\frac{(lat*lon)}{4}}$. Here, $(lat,lon)$ refers to the climate simulator data dimensions. These models are run until convergence.

\end{document}